%% file: main.tex
\documentclass{article}

\usepackage[preprint]{neurips_2024}

\input{preamble}

\title{Active MRI Acquisition with Diffusion Guided Bayesian Experimental Design}

\author{
    Jacopo Iollo$^{*\text{\hspace{0.35em}}1}$
    \And
    Geoffroy Oudoumanessah\thanks{Equal contribution}$^{\text{\quad}1,2,3}$
    \And
    Carole Lartizien$^{2}$
    \And
    Michel Dojat$^{3}$
    \And
    Florence Forbes$^{1}$
}

\begin{document}
\footnotetext[1]{Univ. Grenoble Alpes, Inria, CNRS, Grenoble INP, LJK, 38000 Grenoble, France}
\footnotetext[2]{Univ. Lyon, CNRS, Inserm, INSA Lyon, UCBL, CREATIS, UMR5220, U1294, F‐69621, Villeurbanne, France}
\footnotetext[3]{Univ. Grenoble Alpes, Inserm U1216, CHU Grenoble Alpes, Grenoble Institut des Neurosciences, 38000 Grenoble, France}
\maketitle

\begin{abstract}
A key challenge in maximizing the benefits of Magnetic Resonance Imaging (MRI) in clinical settings is to accelerate acquisition times without significantly degrading image quality. This objective requires a balance between under-sampling the raw $\kv$-space measurements for faster acquisitions and gathering sufficient raw information for  high-fidelity  image reconstruction and analysis tasks. To achieve this balance, we propose to use sequential Bayesian experimental design (BED) to provide an adaptive and task-dependent selection of the most informative measurements. Measurements are sequentially augmented with new samples selected to maximize information gain on a posterior distribution over target images. Selection is performed via a gradient-based optimization of a design parameter that defines a subsampling pattern. In this work, we introduce a new active BED procedure that leverages diffusion-based generative models to handle the high dimensionality of the images and employs stochastic optimization to select among a variety of patterns while meeting the acquisition process constraints and budget. So doing, we show how our setting can optimize, not only standard image reconstruction, but also any associated image analysis task. The versatility and performance of our approach are demonstrated on several MRI acquisitions.
\end{abstract}

\section{Introduction}
Magnetic Resonance Image (MRI) analysis is based on images obtained with reconstruction algorithms. These algorithms process raw measurements acquired by a scanner, which are represented as Fourier coefficients in a so-called $\kv$-space, to produce the images that are generally analyzed, in a so-called $\xv$-space.  The challenge of correctly sampling the $\kv$-space for high-fidelity image reconstructions is a critical aspect of MR research. A high  $\kv$-space sampling rate  increases the patient's time in the scanner, potentially causing discomfort and delaying diagnosis. Conversely, reconstruction quality is directly impacted by $\kv$-space undersampling. Not only does image reconstruction become increasingly challenging as fewer $\kv$-space samples are collected, but sample locations also matter. It has been shown that the $\kv$-space center (lower frequencies) needs to be sampled more densely than the periphery (higher frequencies), see \textit{e.g.} \cite{puy2011,chauffert2014,boyer2019}. Additionally, the target image analysis task (\textit{e.g.} tissue segmentation or anomaly detection), although generally performed from the $\xv$-space image,  is not independent from the acquisition process and can also benefit from an optimized sampling. It follows that there is a strong need for a more integrated approach that jointly optimizes $\kv$-space sampling and image reconstruction quality in $\xv$-space, with respect to the analysis task to be performed. MRI scanners acquire measurements over time, allowing  sequential strategies that adapt subsampling during acquisition. In this work, we propose such a strategy, where the  set of acquired $\kv$-space measurements is gradually enriched with new batches selected so as to maximize information for image reconstruction, and more generally for any associated image analysis task for which training data are available. We assume that a subsampled batch can be fully specified by a  so-called design parameter $\xib \in \Rset^d$ (examples of such $\xib$ are given in Subsection~\ref{sharingan:sec:active}). We propose then to formulate batch selection as an experimental design choice and to use sequential Bayesian experimental design  (BED) to identify the best sequence of batches as determined by the best sequence of design parameters $\xib_1, \xib_2,$ \textit{etc.} Under budget constraints or limited sampling possibilities,  sequential BED allows for the dynamic selection of the next most informative  $\kv$-space measurements over  acquisition time. At each acquisition step, design optimization is performed  within a Bayesian inverse problem, where a posterior distribution is updated in turns and used to identify the most compatible reconstructions and analysis with the set of all subsampled raw measurements so far. Gradient-based stochastic optimization is used to maximize the  expected information gain (EIG), which measures the confidence gained  between two successive posterior distribution updates. To handle the high dimensionality and the intractability of these successive posterior distributions, we leverage a recent BED technique \citep{iollo2025} based on diffusion and show how it can be used to handle very general image recovery task and open the way to more efficient MR image acquisition.

\paragraph{Related work}
Although most traditional acceleration approaches are based on some hand-crafted pre-scan subsampling,  more and more subsampling optimization techniques are becoming available. While advanced acquisition strategies have relied on traditional compressed sensing methods \citep{lustig2008}, even faster and more accurate reconstructions have been obtained with a new generation of deep learning approaches. Generative models or data-driven approaches \citep{chung2022,song2022,wen2023,zhang2018,bangun2024}, particularly diffusion-based models, have been extensively employed to accelerate reconstruction tasks. Diffusion models act as data-based priors by sampling MR images. Conditional diffusions are used to solve the reconstruction inverse problem by using the observed measurements to guide the sampling process. However, to our knowledge, such diffusion models have not been used to select optimal subsampling schemes, with the exception of the work of \cite{nolan2024} which defines an Active Diffusion Subsampling (ADS) approach. Although similar in spirit to our proposal, ADS relies on approximations for tractability, which limits its performance. It can only handle discrete finite design parameters, resulting in very simple subsampling patterns such as straight lines, while it has been shown  that more complex trajectories in the  $\kv$-space are beneficial \textit{e.g.} \citep{boyer2016}. In contrast, several papers, such as \citep{bakker2020,pineda2020,yen2024}, have investigated Reinforcement Learning (RL) for learning adaptive  subsampling schemes. RL and BED share common features, but it has been shown \citep{iollo2024,iollo2025} that BED was outperforming RL in design optimization tasks similar to subsampling optimization. Regarding BED, it has been proposed by \cite{orozco2024} to optimize the masking of the $\kv$-space using normalizing flows to deal with high-dimensional images. However, their experimental evaluation seems preliminary and shows quite high computational costs. In this paper, we rather build on the power of diffusion models to handle more efficiently high dimensional BED settings \citep{iollo2025}. In addition, in contrast to all these previous works, which only address $\xv$-space reconstruction,  we handle  both the $\xv$-space and downstream-task image analysis.
The paper main contributions can be summarized as follows:
\begin{itemize}
 \item A new  procedure is proposed that decomposes MRI acquisition into a number of steps that are dynamically optimized. Each step is seen as an \textit{experiment} in which the next $\kv$-space measurements are selected to maximize the information gain they provide with respect to the target task. It follows an efficient and flexible incremental susbsampling that can be monitored and stopped when a given reconstruction quality is reached.
    \item  The Bayesian inverse problem underlying the procedure is formulated so as to provide knowledge both on reconstructed images and other analysis output such as lesion or tissue segmentations.
    \item High-dimensionality and tractability issues are handled by a careful adaptation of conditional diffusion models to MRI
\end{itemize}

\section{Task-aware accelerated MRI acquisition}

In this section, we first specify the quantities required to formulate traditional MR reconstruction and its acceleration as a Bayesian inverse problem, from observed subsampled measurements. We then extend this inverse problem by augmenting the reconstruction target with an additional analysis task, such as segmentation or anomaly detection, which is also be informed by $\kv$-space measurements.

\begin{figure}[htbp]
    \centering
    \includegraphics[scale=.28]{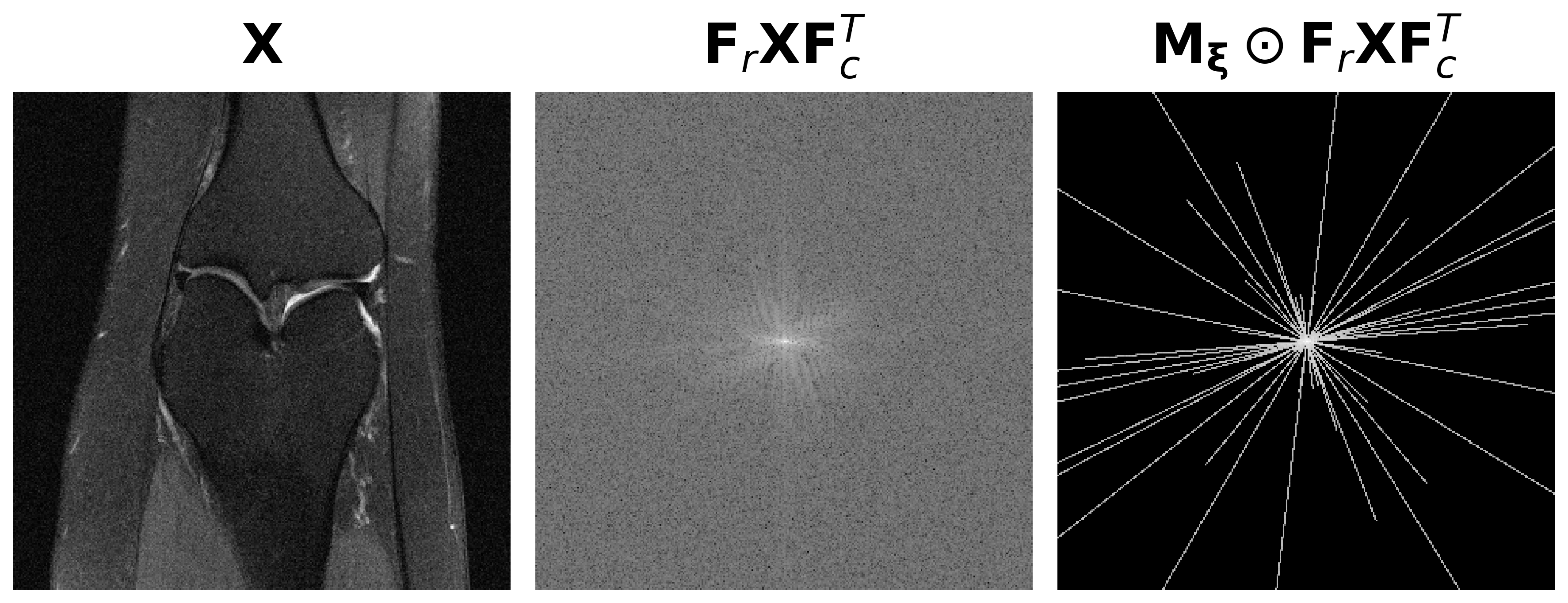}
    \caption{\footnotesize Illustration of the forward model using the fastMRI knee dataset \citep{zbontar2019}: Original knee image $\Xv$ in the $\xv$-space (Left), Fourier Transform of $\Xv$ (Middle) and measurements $\Yv$ actually acquired in the $\kv$-space, for a given subsampling mask $\Mb_\xib$ (Right). The inverse problem consists in recovering an estimation of the left image from the right one. }
    \label{sharingan:fig:forward_model}
\end{figure}

\paragraph{Direct MR image formation model}
In MRI, the produced $\xv$-space images are reconstructed from  $\kv$-space (Fourier space) measurements acquired sequentially from a receiver coil, positioned around the target object,  \textit{e.g.} body tissues. These measurements correspond to the responses of the object to various scanning parameters, \textit{e.g} magnetic field gradients, radio-frequency pulses, etc. \citep{liang2000}. For simplicity, we present the case of a single coil. Multi-coil extensions are similar but are left for future work. Although 3D reconstruction techniques have been studied, \textit{e.g.} by \citet{bangun2024,chaithya2022}, we consider the most common case of reconstruction of 2D images. In practice 3D volumes can be handled as stacks of 2D slices. We denote by $\Xv \in \mathbb{C}^{d_r \times d_c}$ the 2D $d_r \times d_c$ image in the $\xv$-space. In the $\kv$-space, the full-sampled image of the same size is the Fourier transform of $\Xv$ but for acceleration purpose, this image is not fully measured. A $\kv$-space subsampling pattern is represented by a two dimensional $d_r\times d_c$ binary matrice $\Mb_\xib \in \{0,1\}^{d_r \times d_c}$  where elements set to 1 represent the sampled points. In addition, this subsampling operation is assumed to depend on a design parameter $\xib \in \Rset^d$ to be specified in Subsection~\ref{sharingan:sec:active}. Denoting by $\Fb_r$ (resp. $\Fb_c$) the $d_r\times d_r$ (resp. $d_c\times d_c)$) matrix representing the  1D discrete Fourier Transform (DFT) in dimension $d_r$ (resp. $d_c$),  the link between subsampled raw $\kv$-space measurements $\Yv \in \mathbb{C}^{d_r \times d_c}$ and the $\xv$-space image $\Xv$ can be written as
\begin{equation}
    \Yv = \Mb_\xib \odot \Fb_r \Xv \Fb_c^T + \Eb, \label{sharingan:def:mask}
\end{equation}
where $\odot$ denote the element-wise or Hadamard product and $\Eb$ is the complex noise that affects the $\kv$-space measurements. Problem \eqref{sharingan:def:mask} can be equivalently formulated with vectors using the Kronecker product $\otimes$ and $vec (\Ab\Bb\Cb) = (\Cb^T \otimes \Ab) vec(\Bb)$.  The mask $\Mb_\xib$ is converted to a  binary $d_r d_c \times d_r d_c$ diagonal matrix $\Sb_\xib = diag(vec(\Mb_\xib))$ with the elements of vector $vec(\Mb_\xib)$ on the diagonal. Denoting $\xv=vec(\Xv)$, $\yv=vec(\Yv)$ and $\eb=vec(\Eb)$ the vectorized forms of $\Xv$, $\Yv$ and $\Eb$, \eqref{sharingan:def:mask} then  equivalently writes,
\begin{equation}
    \yv = \Sb_\xib (\Fb_c \otimes \Fb_r) \xv + \eb
        = \Ab_\xib \xv + \eb ,
        \label{sharingan:def:forwardvec}
\end{equation}
where  we introduce $\Ab_\xib = \Sb_\xib (\Fb_c \otimes \Fb_r)$ to simplify the notation. We will equivalently use matrix and vector formulations. Considering all images as realizations of random variables, reconstruction can  be cast into a Bayesian inverse problem, where a posterior distribution is estimated and used to identify the most compatible reconstructions  with the subsampled raw measurements, see Figure~\ref{sharingan:fig:forward_model} for an illustration.

\paragraph{Joint MRI reconstruction and analysis}
Standard MR image reconstruction consists of solving a Bayesian inverse problem for the forward model \eqref{sharingan:def:forwardvec}. In our task-aware approach of acquisition, we aim at adding to the image  reconstruction $\Xv$, the recovery of another image representing a downstream analysis of $\Xv$ such as a segmentation. To illustrate this setting, we consider an anomaly segmentation  task and we target an additional binary image $\Zv$ which locates anomalies in $\Xv$. We denote by $\Thetab =(\Xv, \Zv)$ the two images to be recovered from $\Yv$ and then by  $\thetab =(\xv, \zv)$ the concatenation of length $2d_rd_c$ of the two corresponding vectors. The rationale is that in most MRI analysis, only the image magnitude $|\Xv|$ is used, which implies the loss of phase information. We propose to investigate whether $\kv$-space information can be more directly used  for downstream tasks. In practice, there is a clear dependence between $\Xv$ and the downstream objective $\Zv$ but  no clear direct link between $\Zv$ and $\Yv$. We assume that $\Yv$ and $\Zv$ are conditionally independent given $\Xv$. The forward model \eqref{sharingan:def:forwardvec} extends to
\begin{equation}\label{sharingan:eq:forward_model}
    \yv = \overline{\Ab}_\xib \thetab + \eb
\end{equation}
where $\overline{\Ab}_\xib$  is a $d_rd_c \times 2d_rd_c$ matrix whose first $d_rd_c$ columns are those of ${\Ab}_\xib$ and the remaining $d_rd_c$ ones are $\zero$. Note that in  the above formula, seeing $\overline{\Ab}_\xib \thetab$ as the noise free signal, $\eb$ is a complex noise variable that affects the observation of the $\kv$-space \textit{a priori} unknown. Assuming \citep{aja2016} that the noise affects equally all the frequencies and also that it is stationary we can write:
\begin{equation*}
    \eb = \eb_r + j  \eb_i  \; ,\label{sharingan:eq:eta_stationary}
\end{equation*}
where $\eb_r$ and $\eb_i$ are standard Gaussian variables ${\cal N}(\zero, \sigma^2 \Ib)$ standing for the real and imaginary parts of the complex $\eb$. Numerically, we tackle the issue of complex values by implementing any complex variable $\eb$ as a stacked array composed of the real and imaginary part. The  likelihood we consider then writes,
$$p(\yv | \thetab=(\xv, \zv); \xib) = p(\yv | \xv , \xib),$$
which needs to be completed by a prior distribution on $\thetab$, denoted by $p(\thetab) = p(\xv,\zv)$ and assumed to be independent on $\xib$ the subsampling design parameter. The goal and challenge is to produce samples from the posterior distribution below which is generally intractable and involves high dimensional images $\thetab=(\xv,\zv)$,
$$p(\thetab | \yv, \xib) \propto p(\thetab) p(\yv | \xv , \xib).$$
In the next section, we recall how generative diffusion models can be leveraged to sample such posterior distributions.

\section{Posterior sampling with diffusion models}\label{sharingan:sec:diffusion}

\paragraph{\textbf{Data-based samplers with diffusions}} Diffusion-based models are generative models that have greatly extended sampling possibilities, in particular to much higher dimensions, by allowing sampling from distributions only available through a set of examples. Given such a set of $\{\thetab_i=(\xv_i, \zv_i)\}_{i=1:N}$ pairs of $\xv$-space and segmentation maps, which are assumed to be distributed according to some prior distribution $p(\thetab)$, diffusion models are based on the addition of noise to the available samples in such a manner that allows to learn the reverse process that "denoises" the samples. This learned process can then be exploited to generate new samples by denoising easy to simulate random noise samples until we get back to the original data distribution. As an appropriate noising process, in our experiments we ran the Variance Preserving Stochastic Differential Equation (SDE) from \citep{dhariwal2021diffusion}:
\begin{equation} \label{sharingan:eq:vp_sde}
    d\tilde{\thetab}^{(t)} = - \frac{\beta(t)}{2} \tilde{\thetab}^{(t)} dt + \sqrt{\beta(t)} d\tilde{\Bb}_t
\end{equation}
where $\beta(t) > 0$ is a linear noise schedule that controls the amount of noise added at time $t$. Solving SDE \eqref{sharingan:eq:vp_sde} leads to
\begin{equation} \label{sharingan:eq:vp_noisy_sample_2}
\small
    \tilde{\thetab}^{(t)} \!= \!\sqrt{\bar{\alpha}_t} \tilde{\thetab}^{(0)}\! + \!\sqrt{1-\bar{\alpha}_t} \epsilonb \; \text{with} \;\bar{\alpha}_t = \exp(-\int_0^t \beta(s) ds)
\end{equation}
and where $\epsilonb \sim \mathcal{N}(\zero,\Ib)$ is a standard Gaussian random variable. Iterating (\ref{sharingan:eq:vp_noisy_sample_2}) for some large  time $T$, samples $\tilde{\thetab}^{(0)}$ from the prior $p(\thetab)$ are gradually transformed  to samples closed to a standard Gaussian distribution. The reverse denoising process can then be written as the reverse of the diffusion process (\ref{sharingan:eq:vp_sde}), which as stated by
\cite{anderson1982reverse} is:
\begin{equation} \label{sharingan:eq:vp_reverse_sde}
\small
    d\thetab^{(t)}\! = \!\left[-\frac{\beta(t)}{2} \thetab^{(t)}\! - \!\beta(t) \nabla_\thetab \log p_t(\thetab^{(t)})\!\right] \!dt \!+ \!\sqrt{\beta(t)} d{\Bb}_t
\end{equation}
where $p_t$ is the distribution of $\tilde{\thetab}^{(t)}$ from \eqref{sharingan:eq:vp_sde}. Solving this reverse SDE, the distribution of $\thetab^{(T)}$ is closed to $p(\thetab)$ for large $T$. The score $\nabla_\thetab \log p_t(\thetab^{(t)})$ of the noisy data distribution at time $t$  is intractable and is then estimated by learning a neural network $s_\phi(\thetab,t)$ with parameters $\phi$. We use score matching \citep{hyvarinen2005,song2021denoising} to train $s_\phi$. Details are available in Supplementary~\ref{sharingan:sec:training}. Once the neural network $s_\phi$ has been trained, it can be used to generate new samples approximately distributed as the target prior by running a numerical scheme on the reverse SDE \eqref{sharingan:eq:vp_reverse_sde}.

\paragraph{Posterior sampling with conditional diffusions}
To solve the inverse problem of recovering MR images $\thetab$ from $\kv$-space measurements $\yv$ resulting from \eqref{sharingan:eq:forward_model}, we need to  produce samples from some conditional distribution $p(\thetab|\yv,\xib)$. When using diffusion models, numerous solutions have been investigated as mentioned in a  recent review \citep{daras2024}. Sampling from the conditional distribution $p(\thetab|\yv, \xib)$ can be done by running the reverse diffusion process on the following conditional SDE,
\begin{equation}\label{sharingan:eq:vp_reverse_sde_conditional}
\small
    d\thetab^{(t)}\!\! = \!\!\left[ - \frac{\beta(t)}{2} \thetab^{(t)}\! \!-\! \beta(t)\! \nabla_\thetab \!\log \!p_t(\thetab^{(t)}\!| \yv, \xib)\! \right]\!\!dt \!+ \!\!\sqrt{\beta(t)} d\Bb_t
\end{equation}
with the prior score $\nabla_\thetab \log p_t(\thetab^{(t)})$ replaced by the conditional score $\nabla_\thetab \log p_t(\thetab^{(t)} | \yv, \xib)$. The main challenge of conditional diffusions is to generate samples from $p(\thetab|\yv, \xib)$ without retraining a new neural network for the new conditional score. This score is linked to the prior score via
$$\nabla_\thetab \log p_t(\thetab^{(t)}| \yv, \xib)= \nabla_\thetab \log p_t(\thetab^{(t)}) + \nabla_\thetab \log p_t(\yv| \thetab^{(t)},\xib),$$
but the additional term involves
$$p_t(\yv| \thetab^{(t)},\xib) = \int p_t(\yv| \thetab^{(0)},\xib) p(\thetab^{(0)}| \thetab^{(t)}) d\thetab^{(0)},$$
which is difficult to evaluate as $p(\thetab^{(0)}| \thetab^{(t)})$ is only implicitly defined through the diffusion model. Conditional diffusion implementations then mainly differ in the way they approximate the conditional score. An interesting solution with good theoretical guarantees is proposed by \cite{boys2024}. It approximates  $p(\thetab^{(0)}| \thetab^{(t)})$ with a Gaussian using Tweedie's first and second moments formulas \citep{efron2011} but is too computationally costly to implement with large dimensional  images. Instead we use a simpler variant, Diffusion Posterior Sampling (DPS) proposed by \cite{chung2022}, which approximates $p(\thetab^{(0)}| \thetab^{(t)})$ by a Dirac mass at its expectation $\mathbb{E}[\thetab^{(0)} | \thetab^{(t)}]$ and  uses only Tweedie's first moment to compute $ \Exp(\thetab^{(0)} | \thetab^{(t)}) ={\cal T}(\thetab^{(t)})$ with
\begin{equation}
    {\cal T}(\thetab^{(t)}) = \frac{\thetab^{(t)} + (1 - \bar{\alpha}_t) \nabla_\thetab \log p_t(\thetab^{(t)})}{\sqrt{\bar{\alpha}_t}}
    \label{sharingan:def:tweedie}
\end{equation}
Denoting $\hat{\thetab}_0^{(t)}= {\cal T}(\thetab^{(t)}) $, $p_t(\yv| \thetab^{(t)},\xib) \approx p_t(\yv| \hat{\thetab}_0^{(t)},\xib)$, which is a Gaussian distribution, as assumed in \eqref{sharingan:eq:forward_model}, so that $\nabla_\thetab \log p_t(\yv|\thetab^{(t)}, \xib)$ is approximated by
\begin{equation}\label{sharingan:eq:DPSbis}
    \frac{1}{\sigma^2 \sqrt{\bar{\alpha}_t}} \left( \Ib + (1-\bar{\alpha}_t)\nabla^2_\thetab \log p_t(\thetab^{(t)}) \right)^{H} \; \overline{\Ab}_\xib^H \left( \yv - \overline{\Ab}_\xib\hat{\thetab}_0^{(t)} \right)
\end{equation}

\section{Active MRI acquisition with sequential experimental design}
\label{sharingan:sec:active}
In MRI, $\kv$-space measurements are acquired sequentially, but this acquisition has to follow some physically feasible trajectories. Typically, although often considered in previous papers, random pixel selection is not easily implementable on scanners.

\begin{figure}[htbp]
    \centering
    \includegraphics[scale=.165]{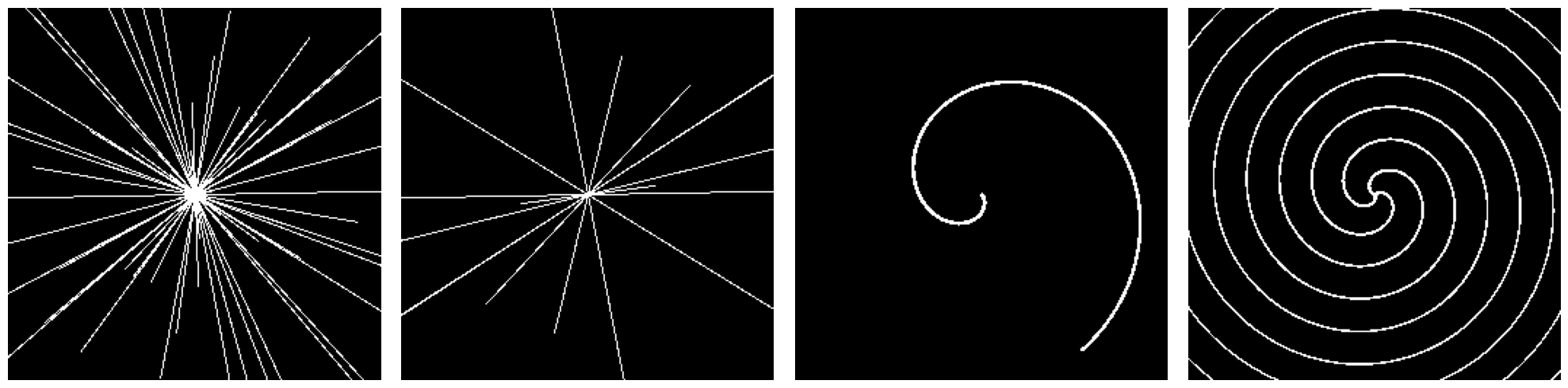}
    \caption{\footnotesize Examples of sampling patterns for data acquisition: (from left to right) radial sampling with 40 lines, with 10 lines, spiral sampling  with 1 spiral and 3 interleaved spirals.}
    \label{sharingan:fig:sampling_patterns}
\end{figure}

\paragraph{Subsampling patterns}
In this work, we consider only feasible trajectories, which currently include so-called Cartesian, spiral and radial schemes \citep{zbontar2019}. Figure~\ref{sharingan:fig:sampling_patterns} shows an illustration of these schemes. Among these 3 schemes, different acquisition trajectories are possible and have to be specified. The subsampling strategy of the $\kv$-space is assumed to depend on a parameter $\xib$, which we can control and is thus referred to as a design parameter. $\xib$ is assumed to determine a subsampling  trajectory in the $\kv$-space. In the Cartesian case, $\xib$ can represent the vertical or horizontal  coordinates that define the chosen lines.  In the radial case, $\xib$ determines the angles and lengths of the chosen radial lines. In the spiral case $\xib$ is a 3-dimensional vector  \citep{glover2012} for a single spiral and  higher dimensional if it represents several of them.

\paragraph{Bayesian optimal experimental design} The issue of selecting $\kv$-space measurements to optimize the trade-off between acquisition speed and reconstruction and detection quality can be formulated as a BED objective.
Bayesian optimal experimental design (BED) \citep{chaloner1995,sebastiani2000,amzal2006} has  recently gained new interest with the use of machine learning techniques, see \cite{rainforth2024,huan2024} for  recent reviews. The most common approach consists of  maximizing the so-called expected information gain (EIG), which  is a mutual information criterion that accounts for information via the Shannon's entropy.

The EIG, denoted below by $I$, admits several equivalent expressions, see \textit{e.g.} \citet{foster2019}. It can be written as the expected loss in entropy when accounting for an observation $\yv$ at $\xib$ \eqref{sharingan:def:I} or as a mutual information or expected Kullback-Leibler (KL) divergence \eqref{sharingan:def:EIGpost},

\begin{align}
I(\xib) & =  \Exp_{p(\yv | \xib)}[\text{H}(p(\thetab)) - \text{H}(p(\thetab | \yv, \xib)]  \label{sharingan:def:I} \\
&=  \Exp_{p(\yv | \xib)}\left[\text{KL}(p(\thetab | \yv, \xib), p(\thetab))\right]   \; , \label{sharingan:def:EIGpost}
\end{align}
where $\Exp_{p(\cdot)}[\cdot]$ or $\Exp_{p}[\cdot]$ denotes the expectation with respect to $p$ and  $\text{H}(p(\thetab))= -\Exp_{p(\thetab)}[\log p(\thetab)]$ is the entropy of $p$. In BED, we  look for $\xib^*$ satisfying
\begin{equation}
\xib^*  \in \arg\max_{\xib \in \Rset^d} I(\xib)
\;. \label{sharingan:def:xistar}
\end{equation}
The challenge in EIG-based BED is that both the EIG and its gradient with respect to $\xib$ are doubly intractable. In addition, most  BED approaches assume that the  prior is available in closed-form. In this work, the relevant prior is only available through samples which are in high dimension (images). To deal with this case, we use the diffusion-based generative models presented in the previous Section,  using a setting similar to the one recently introduced by \cite{iollo2025}.

\begin{minipage}{\linewidth}
\begin{algorithm}[H]
\footnotesize
\SetAlgoLined
\DontPrintSemicolon
\caption{Sequential Bayesian Experimental Design}
\label{sharingan:algo:sequential}

\textbf{Initialisation:}  $\xib_0 \in \Rset^d$, $D_0 = \emptyset$\;
\For{$k=1$ \KwTo $K$ \textnormal{\small(sequential experiments)}}{
    $\thetab_{k-1}\sim p(\thetab|D_{k-1}),\;\xib_k^* $ \hfill \textnormal{\small(Joint Sampling-Optimization )}\;
    $\yv_k \sim p(\yv|\thetab,\xib_k^*)$ \hfill \textnormal{\small(Run Experiment)}\;
    $D_k \leftarrow D_{k-1} \cup \{(\xib_k^*, \yv_k)\}$ \hfill \textnormal{\small(Update Dataset)}\;
}
\Return{\textnormal{\small Optimal designs, posterior samples} $\{\xib_k^*,\thetab_{k-1}\}_{k=1}^K$} \;
\end{algorithm}
\end{minipage}

\paragraph{Sequential design for dynamic subsampling optimization}
Solving  optimization  (\ref{sharingan:def:xistar}) is a \textit{static}  or \textit{one-step} design problem. A single $\xib$ or multiple $\{\xib_1, \cdot, \xib_K\}$ are selected prior to any observation, measurements $\{\yv_1, \cdot, \yv_K\}$ are made for these design parameters and the process is stopped. In MRI, this situation corresponds to the optimization of a pre-scan sampling of the $\kv$-space.  In static design, the selected designs depend only on the model, no feedback is possible from the measurements actually made. In contrast, in sequential or iterated design, $K$ experiments or subsampling patterns are planned sequentially to construct an adaptive strategy, meaning that for the k$^{th}$ experiment, the best $\xib_k$ is selected taking into account the previous design parameters and associated $\kv$-space measurements $\Db_{k-1} =\{(\yv_1,\xib_1), \cdot , (\yv_{k-1}, \xib_{k-1})\}$.  Then, a new set  $\yv_k$ of $\kv$-space measurements are included according to the pattern defined by $\xib_k$ and $\Db_{k}$ is updated into $\Db_k = \Db_{k-1} \cup (\yv_k, \xib_k)$. In this work, we limit ourselves to a greedy approach, replacing in (\ref{sharingan:def:I}) or (\ref{sharingan:def:EIGpost}) the prior $p(\thetab)$ by our current belief on $\thetab$, namely $p(\thetab | \Db_{k-1})  = p(\thetab | \yv_1, \xib_1, \cdot ,  \yv_{k-1},  \xib_{k-1})$, and to solve iteratively for
\begin{align}
\xib_k^* &\in \arg\max_{\xib \in \Rset^d} I_k(\xib), \label{sharingan:def:seqd}
\end{align}
\vspace{-.5cm}
\begin{align}
\mbox{where } & I_k(\xib)  =  \Exp_{}[H\!(p(\thetab | \Db_{k-1}))\!- \!H\!(p(\thetab  | \Yv, \xib, \!\Db_{k-1}))] \nonumber \\
  &=  \Exp_{}\left[\text{KL}(p(\thetab  | \Yv, \xib, \!\Db_{k-1}), p(\thetab | \Db_{k-1}))\right]
    \label{sharingan:def:Ik}
    \end{align}
    and $\Exp$  is  with respect to $p(\yv | \xib, \Db_{k-1})$.
Observations are   assumed conditionally independent so that
$\!p(\thetab | \Db_k) \!\propto \!p(\thetab) \prod_{i=1}^k\! p(\yv_i | \thetab, \xib_i)\!$ which also leads to
\begin{align}
p(\thetab | \Db_k) \propto p(\thetab | \Db_{k-1}) \; p(\yv_k | \thetab, \xib_k) \; . \label{sharingan:eq:ci}
\end{align}
Non-greedy approaches exist, using for instance reinforcement learning principles \citep{blau2022,foster2021} but with another layer of complexity and performance that are not always superior, see  \cite{blau2022,iollo2024,iollo2025}.

In most BED procedures, solving \eqref{sharingan:def:seqd} is performed via a stochastic gradient approach, which requires an estimate of the gradient $\nabla_\xib I_k$. The optimization is intrinsically linked to sampling from the posterior distribution \eqref{sharingan:eq:ci}. Posterior samples are not only the target outcome of Bayesian resolution of inverse problems but in a lot of approaches they are also used in Monte Carlo estimations of the intractable EIG gradients.  We adopt this joint sampling-optimization approach (Algorithm~\ref{sharingan:algo:sequential}) following the implementation proposed by \cite{iollo2025} that can handle high dimensional images without exploding computational cost.

\paragraph{Joint Sampling-Optimization Procedure}
\cite{iollo2025} manage to handle computational cost by proposing a sampling-as-optimization point of view to unify the optimization of the design \eqref{sharingan:def:seqd} and the sampling of the particles \eqref{sharingan:eq:ci} needed for sequential design. The procedure relies on an expression of the EIG gradient as a function $\Gamma$ of the design $\xib$, the joint and a so-called pooled posterior distributions, $\nabla_\xib \Ib(\xib)= \Gamma(p(\yv,\thetab|\Db_{k-1}), q(\thetab|\xib),\xib)$. At each joint sampling-optimization step $t$, the EIG gradient can be estimated with samples $\{(\thetab_i^{(t)}, \yv_i^{(t)})\}_{i=1:N}$, $\{\thetab_j^{'(t)}\}_{j=1:M}$ from these distributions. The conditional diffusion models introduced in Subsection~\ref{sharingan:sec:diffusion} provide values for $\thetab_i^{(t)}$ and $\thetab_j^{'(t)}$.

However, for small $t$, these samples are still far from the target distributions and would lead to low-informative gradients. Tweedie's formula \eqref{sharingan:def:tweedie} is then applied to $\thetab^{(t)}_i$  to mitigate this problem leading to new samples  $\hat{\thetab}^{(t)}_{0i}$ and  $\yv^{(t)}_i$ are  simulated, using \eqref{sharingan:def:forwardvec} with $\hat{\thetab}^{(t)}_{0i}$. Samples $\thetab_j^{'(t)}$ are simulated from the resulting pooled posterior and Tweedie's formula is applied again to get $\thetab_{0j}^{'(t)}$. EIG gradients are  then estimated using $\{(\hat{\thetab}_{0i}^{(t)}, \yv_i^{(t)})\}_{i=1:N}, \{\hat{\thetab}^{'(t)}_{0j}\}_{j=1:M}$. It follows a joint sampling-optimization that alternates a sampler step for $p_t(\thetab|\Db_{k-1})$ and $q_t(\thetab|\xib)$ and a design $\xib$ optimization step  with a gradient  step. The procedure is summarized  in Algorithm~\ref{sharingan:algo:joint}, where  $\Sigma_t^\theta$ and $\Sigma_t^{\theta'}$  denote  diffusion-based sampling operators. Details can be found in Supplementary~\ref{sharingan:app:codiff}.

 \begin{algorithm}[H]
\SetAlgoLined
\KwResult{Optimal design $\xib^*$}
\textbf{ Initialisation:} $\xib_0\!\in\! \Rset^d$,
\\
\For{t=0:T-1 \mbox{\!(sampling-optimization loop)} }{
    $\thetab_i^{(t+1)} = \Sigma_t^{\thetab}(\thetab_i^{(t)},\xib_t)$ (i=1:N)\\
    $\hat{\thetab}_{0i}^{(t+1)} = {\cal T}(\thetab_{i}^{(t+1)})$  (Tweedie prediction)\\
    $\yv_i^{(t+1)} \sim p(\yv | \hat{\thetab}_{0i}^{(t+1)})$ \\
   ${\thetab}_j^{'(t+1)} = \Sigma_t^{\thetab'}({\thetab}_j^{'(t)},\xib_t, \{\yv_i^{(t)}\}_{i=1:N})$ (j=1:M) \\
   $\hat{\thetab}_{0i}^{'(t+1)} = {\cal T}(\thetab_{i}^{'(t+1)})$  (Tweedie prediction)\\
   Compute \\
   $\quad \nabla_\xib \!I(\xib_t)\! \approx\!\hat{\Gamma}(\{(\hat{\thetab}_{0i}^{(t)}, \yv_i^{(t)})\}_i, \{\hat{\thetab}^{'(t+1)}_{0j}\}_j,\xib_t)$\\
    Update $\xib_t$ with  SGD or another optimizer
}
\Return{ $\xib_T$}\;
\caption{Contrastive Diffusions (CoDiff) \label{sharingan:algo:joint}}
\end{algorithm}

\section{Experiments}
We first demonstrate, using an open dataset of knee images fastMRI \citep{zbontar2019} the performance of our method in terms of accelerated reconstruction. We then show that this acceleration can also be coupled with a downstream task. More specifically, we consider a segmentation task of white matter hyper-intensities, performed on brain images from the WMH dataset \citep{WMH2022}.

\subsection{Experiment set up and evaluation metrics}
For comparison with previous work, we consider the reconstruction of  MR images from a total budget of 25\% of the $\kv$-space measurements. In our BED procedure, we consider $K=20$ experiments at the end of which the $\kv$-space should be sampled at about 25\%. When considering radial patterns, this amounts of choosing at each experiment, the angle of 15 lines, resulting in $\xib_k \in \Rset^{15}$, for $k=1:K$.

To evaluate reconstruction quality, we use the structural similarity index measure (SSIM) of \citet{wang2004} and the peak signal-to-noise ratio (PSNR). The SSIM compares the magnitude of the ground truth target image and the reconstructed image. It is computed using a window of $7\times 7$, $k_1=0.01$ and $k_2=0.03$, as set in the fastMRI challenge's implementation. To assess segmentation results, we use the Dice similarity metric \cite{dice1945}. This metric measures the overlap between a segmentation result and the gold standard. By denoting by $TP_c$ the number of true positives for class $c$, $FP_c$ the number of false positives and $FN_c$ the number of false negatives the Dice metric is given by $Dice_c= \frac{2TP_c} {2TP_c+FN_c+FP_c}$ and takes its value in $[0, 1]$ where 1 represents the perfect agreement.

\subsection{Datasets}
\paragraph{Comparison with other reconstruction methods}
To our knowledge, no previous work has addressed both reconstruction and a downstream task simultaneously. Comparison is then only possible on reconstructions. We test our subsampling strategy on a dataset consisting of MR images of the knee from the fastMRI benchmark \citep{zbontar2019}. The fastMRI dataset is specially appropriate for reconstruction evaluation, providing full-sampled $\kv$-space images. For a fair comparison, we follow the same protocol as \citet{nolan2024}. Regarding reconstruction, we also provide results for the brain images in fastMRI and BRATS \citep{zbontar2019} data sets.

\paragraph{Segmentation}
Regarding the downstream task, we focus on a segmentation task. For such an evaluation, note that in common open data sets, only magnitudes of the $\xv$-space  images are available and that these images have been acquired with an already under-sampled $\kv$-space. Full $\kv$-space and  phase information is typically not available. Although we can still apply and show the performance of our method in this case, current available data does not allow to fully assess its potential.

\subsection{Results}
For comparison, we report in Table \ref{sharingan:tab:ssim_scores} the results in Table 2 of \cite{nolan2024}, which shows mean SSIM reconstruction scores for the fastMRI knee data set and a 25\% $\kv$-space, which is considered as a 4 times acceleration. A number of existing methods are compared, namely PG-MRI \citep{bakker2020}, LOUPE \citep{bahadir2020}, SeqMRI \citep{yin2021}, ADS \citep{nolan2024}, and  a baseline referred to as fixed-masked Diffusion Posterior Sampling (DPS). Fixed-mask DPS uses a pre-scan mask as used by \citet{zbontar2019} and \citet{nolan2024}. In addition, we report the result with a strategy referred to as Random that chooses 15 radial lines at random at each experiments.

\begin{figure}[h!]
    \centering
    \hspace*{-1cm}
    \includegraphics[width=\textwidth]{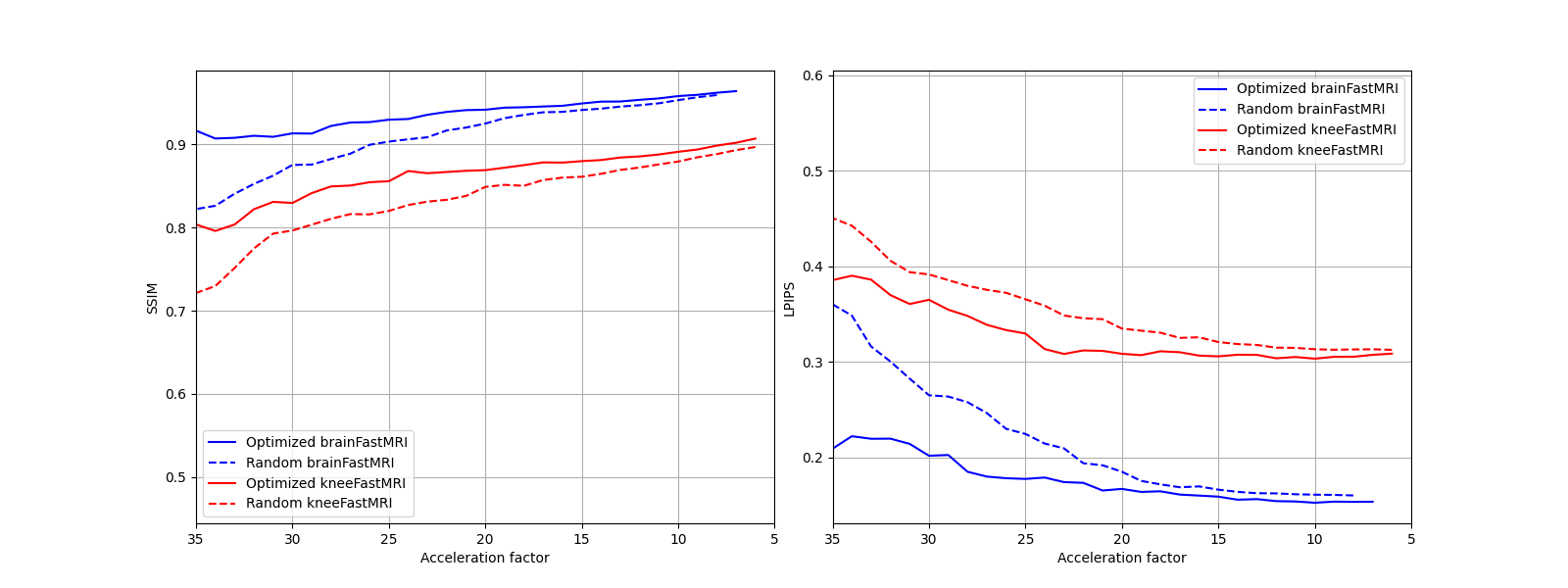}
    \caption{\footnotesize Median SSIM and LPIPS over various data sets wrt the acceleration factor (100 / (\% of $\kv$-space sampled) for radial sampling}
    \label{sharingan:fig:metrics}
\end{figure}

Figure~\ref{sharingan:fig:metrics} further showcases the strength of our approach. Even at a 25 times acceleration (only 4\% $\kv$-space sampling), it achieves SSIM scores between 80 and 90, highlighting its ability to maintain quality with minimal data. The figure also tracks the median SSIM and PSNR as the acceleration factor—defined as $100/k_p$, where $k_p$ is the percentage of sampled $\kv$-space varies. As expected, both metrics improve with increased sampling, surpassing 90 for SSIM and 20 for PSNR at 25\% $\kv$-space sampling. The comparison with random measurements also acts as a reminder that the higher the $\kv$-space is sampled, the less interesting it becomes to optimize the design $\xib$ as for low acceleration factor, random experiments are as effective as optimized ones.

\begin{figure}[h!]
    \centering
    \includegraphics[width=\textwidth]{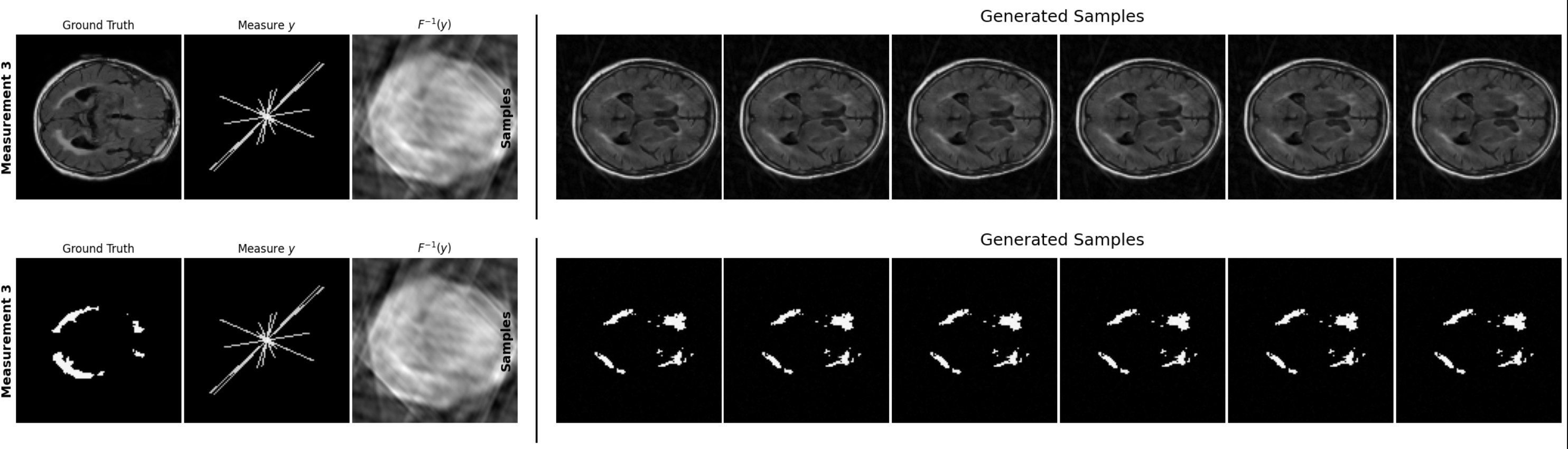}
    \caption{\footnotesize Example of simultaneous reconstruction of the image in the $\xv$-space and its corresponding segmentation map of anomalies}
    \label{sharingan:fig:segment}
\end{figure}

Additionally, Figure~\ref{sharingan:fig:segment} illustrates joint $\xv$-space and segmentation map reconstruction for WMH brain images. It demonstrates that while a precise $\xv$-space reconstruction demands an extensive $\kv$-space sampling, targeted tasks like segmentation can achieve reliable results with far less data, making BED an effective tool for accelerating the reconstruction of the segmentation pattern. Intuitively, to reconstruct the image with very high fidelity in all the details, and to achieve high metrics like SSIM or PSNR, a high percentage of the $\kv$-space has to be sampled. However, even without capturing every nuance, coarser information like segmentation patterns can still be accurately inferred from limited samples.

The sequential framework further allows adaptive sampling adjustments to meet desired quality, and our subsampling patterns align with findings that low $\kv$-space frequencies are key for reconstruction, while higher frequencies enhance tasks like anomaly detection.

\begin{figure*}[h!]
    \centering
        \centering
    \includegraphics[width=\textwidth]{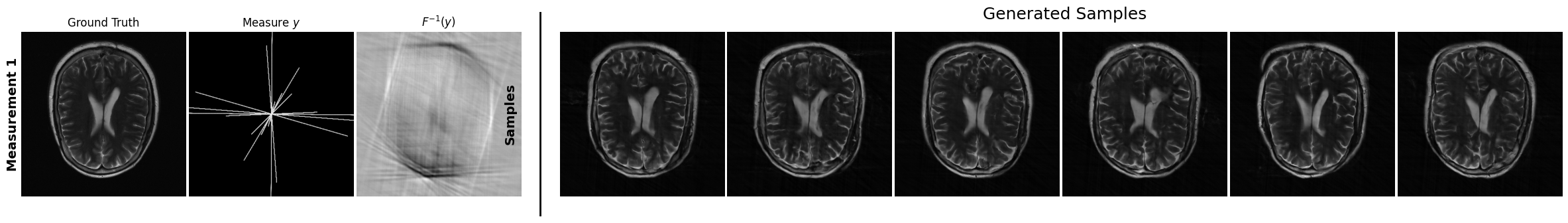}
    \includegraphics[width=\textwidth]{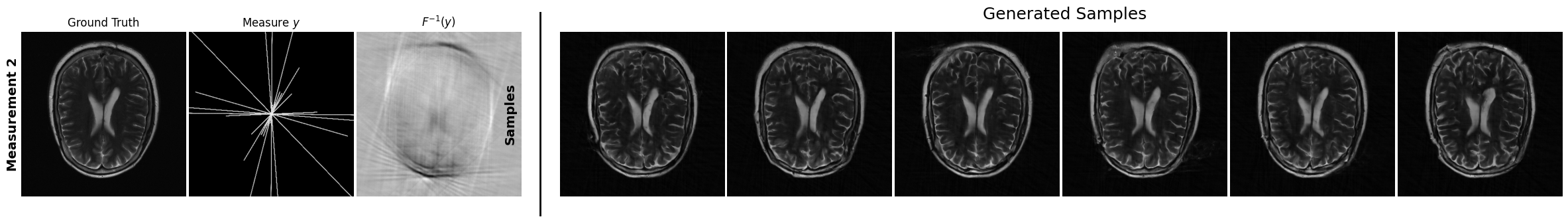}
    \includegraphics[width=\textwidth]{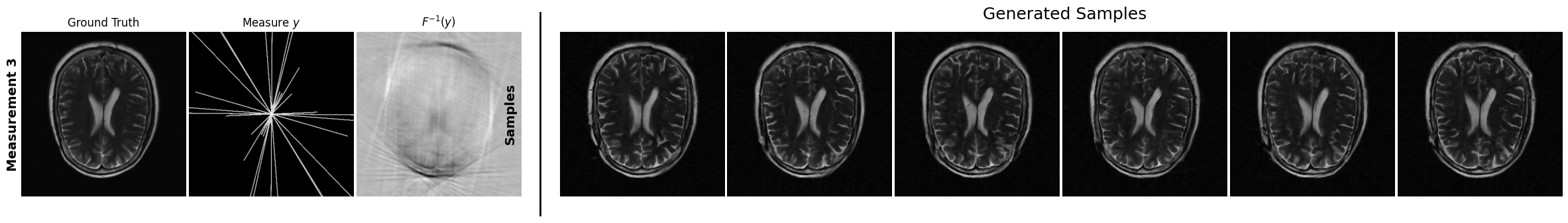}
    \includegraphics[width=\textwidth]{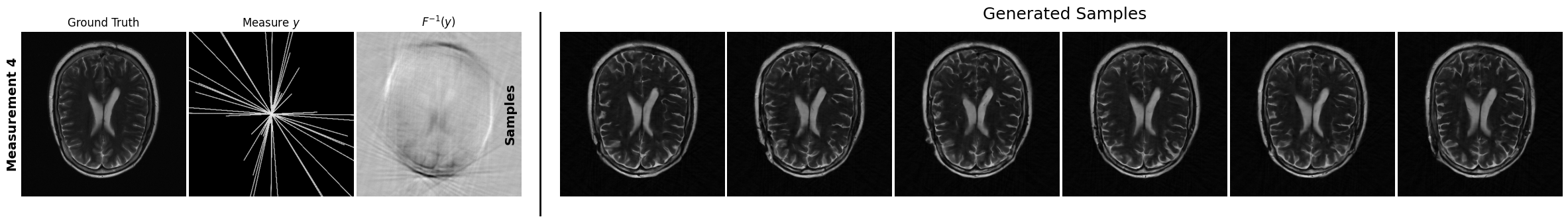}
    \includegraphics[width=\textwidth]{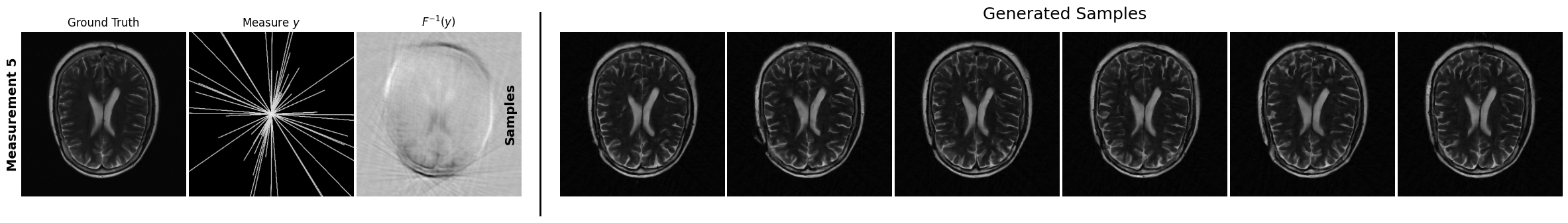}
    \caption{\footnotesize  Sequence of experiments, measures, direct Fourier inversion and samples from posterior generated by our procedure. It illustrates how the data-based prior completes the missing information in the direct Fourier inversion to propose plausible posterior samples}
    \label{sharingan:fig:}
\end{figure*}

\begin{table}[h!]
\centering
\begin{tabular}{cccc}
\toprule
\textbf{Iteration} & \textbf{PSNR} ($\uparrow$) & \textbf{SSIM} ($\uparrow$) & \textbf{$\kv$-space \%}  \\
\midrule
1 & 16.0957 & 0.7890 & 3.3659 \\
3 & 18.5530 & 0.8743 & 6.1733 \\
5 & 19.5297 & 0.8912 & 9.1834 \\
7 & 20.1386 & 0.9021 & 12.3456 \\
9 & 20.6784 & 0.9104 & 15.6789 \\
\bottomrule
\end{tabular}
\caption{Metrics for selected iterations of our BED procedure applied to the brain fastMRI data set. Higher values indicate better performance.}
\label{sharingan:tab:metrics_selected}
\end{table}

\begin{table}[h!]
\centering
\begin{tabular}{cc}
\toprule
\textbf{Method} & \textbf{SSIM} ($\uparrow$) \\
\midrule
 PG-MRI \citep{bakker2020} & 87.97 \\
 LOUPE \citep{bahadir2020} & 89.52 \\
 Fixed-mask DPS \citep{nolan2024} & 90.13 \\
 SeqMRI \citep{yin2021} & 91.08 \\
 ADS  \citep{nolan2024} & 91.26 \\
 Random  &  90.2 \\
  Ours & \textbf{91.34} \\
\bottomrule
\end{tabular}
\caption{Reconstruction quality: mean SSIM for the fastMRI knee test set with a 25\% subsampling of the $\kv$-space.}
\label{sharingan:tab:ssim_scores}
\end{table}

\section{Conclusion and future work}
In contrast to a number of deep learning approaches, which generate only single reconstructed images, we considered the issue of MRI acquisition as a Bayesian inverse problem. This requires sampling from a posterior distribution but provides more information, exploitable to run Bayesian Experimental Design, and useful in particular for downstream inference tasks. To do this, we extended conditional diffusion model to sample from the MRI inverse problem and to tackle a Bayesian experimental design optimization. We illustrated on several challenging MRI tasks that our method outperformed similar existing approaches and was very efficient in capturing information in the $\kv$-space for an optimized acceleration/output quality balance. Furthermore, it is particularly adapted to reconstruct coarser information like segmentation patterns that do not necessitate all the fine grained information that is needed to accurately refine an MRI scan.\\

\paragraph{Limitations} Regarding the first objective of finding a balance between acquisition acceleration and image quality, our current procedure is described for a single coil acquisition setup and 2D images. Generalization to multiple coils is important as it allows to benefit from the acceleration of parallel imaging, which is now available in most scanners. Such an extension should not be problematic following a multiple coil modelling similar to \textit{e.g.} that of \citet{wen2023}. Another important extension is to handle 3D images, as dealing with stacks of 2D slices may underestimate the dependence between slices. The recent work of \citet{bangun2024} on 3D diffusion models could be a good starting point to adapt our BED framework. Regarding, the data-driven and task-oriented aspects of our modelling, part of the procedure performance depends on the quality of the learned priors. When using diffusions, this requires the availability of large enough learning data sets. In particular, when targeting a downstream task, pairs of ground truth of reconstructed and segmented images are necessary. For task-aware procedures, a prerequisite is the availability of training data pairs $(\xv,\zv)$ in quantity large enough to learn a diffusion prior.

\paragraph{Future work} In addition to multiple coil and 3D settings, it would be interesting to investigate other subsampling schemes, in particular as suggested by \cite{lazarus2019,boyer2016}.

\bibliographystyle{plainnat}
\bibliography{main}

\newpage
\appendix

\section{Training }\label{sharingan:sec:training}
Training was done by minimizing the following score matching loss:

\begin{equation} \label{sharingan:eq:score_matching}
     \mathcal{L}(\phi) = \Exp_{p_t(\thetab)} \left[ || s_\phi(\thetab,t) - \nabla_\thetab \log p_t(\thetab) ||^2 \right] \; .
\end{equation}
which in practice takes the form:
\begin{equation} \label{sharingan:eq:score_matching_song}
    \Exp_{t\sim U[0,T]} \Exp_{p_0(\thetab^{(0)})} \Exp_{p_t(\thetab | \thetab^{(0)})} \left[ \lambda(t) || s_\phi(\thetab,t) - \nabla_\thetab \log p_t(\thetab|\thetab^{(0)}) ||^2 \right]
\end{equation}

\begin{table}[h]
\small
\centering
\renewcommand{\arraystretch}{1.1}
\begin{tabular}{llccccc}
\hline
\textbf{Category} & \textbf{Parameter} & \textbf{BRATS} & \textbf{WMH} & \textbf{fastMRI} & \textbf{brainFM} & \textbf{kneeFM} \\
\hline
\multirow{7}{*}{Training} & Epochs & 4000 & 4000 & 4000 & 4000 & 4000 \\
& Train Ratio & 0.8 & 0.8 & 0.8 & 0.8 & 0.8 \\
& Batch Size & 32 & 32 & 32 & 32 & 32 \\
& Time Samples & 32 & 32 & 32 & 32 & 32 \\
& Learning Rate & 2e-4 & 2e-4 & 2e-4 & 2e-4 & 2e-4 \\
\hline
& Optimization & \multicolumn{5}{c}{Adam} \\
& Schedule & \multicolumn{5}{c}{Cosine decay w/ warmup} \\
& Decay Steps & \multicolumn{5}{c}{95\% of total steps} \\
& Final LR & \multicolumn{5}{c}{1\% of initial LR} \\
& Grad Clip & \multicolumn{5}{c}{Global norm = 1.0} \\
& EMA Rate & \multicolumn{5}{c}{0.99} \\
\hline
\multirow{4}{*}{SDE} & $\beta_{min}$ & 0.02 & 0.02 & 0.02 & 0.02 & 0.02 \\
& $\beta_{max}$ & 5.0 & 5.0 & 5.0 & 5.0 & 5.0 \\
& $t_0$ & 0.0 & 0.0 & 0.0 & 0.0 & 0.0 \\
& $t_f$ & 2.0 & 2.0 & 2.0 & 2.0 & 2.0 \\
\hline
\multirow{4}{*}{UNet} & Emb Dim & 128 & 64 & 64 & 64 & 64 \\
& Upsampling & p\_shuffle & p\_shuffle & p\_shuffle & p\_shuffle & p\_shuffle \\
& Dim Mults & [1,2,4,8] & [1,2,4] & [1,2,4,8,8] & [1,2,4,8,8] & [1,2,4,8,8] \\
& dt\_emb & 0.002 & 0.002 & 0.002 & 0.002 & 0.002 \\
\hline
\multirow{2}{*}{Mask} & Type & radial & radial & radial & radial & radial \\
& Num Lines & 5 & 5 & 5 & 5 & 5 \\
\hline
Task & & seg & seg & recon & recon & recon \\
\hline
\end{tabular}
\caption{Training and optimization parameters across different MRI datasets}
\label{sharingan:tab:training_params}
\end{table}

\subsection{Hardware details}
All experiments were run on NVIDIA A100 GPUs 80GB. Training was split over 8 GPUs with a batch size of 64 per GPU for faster training times and to avoid memory issues.

\section{Joint Sampling-Optimization}\label{sharingan:app:codiff}

\paragraph{\textbf{The Pooled Posterior Distribution}}

Given a candidate design \(\xib\), suppose we have access to a set of simulated joint samples
\[
\{(\thetab_i, y_i)\}_{i=1}^{N} \sim p_\xi(\thetab,y)=p(\thetab)\,p(y\mid \thetab,\xi).
\]
We define the \emph{pooled posterior} as a logarithmic pooling of the individual posteriors:
\begin{equation}\label{sharingan:eq:pooled_posterior}
  q_{\xi, \rho}(\thetab) \propto \exp\Bigl( \mathbb{E}_{\rho}\bigl[ \log p(\thetab \mid Y, \xi) \bigr] \Bigr),
\end{equation}
where \(\rho\) is a probability measure on the observation space \(Y\). In the common case where
\[
\rho(y) = \sum_{i=1}^{N} \nu_i \, \delta_{y_i}(y) \quad \text{with} \quad \sum_{i=1}^{N} \nu_i = 1,
\]
the pooled posterior simplifies to
\begin{equation}\label{sharingan:eq:pooled_posterior_discrete}
  q_{\xib,N}(\thetab) \propto \prod_{i=1}^{N} \bigl[ p(\thetab \mid y_i, \xib) \bigr]^{\nu_i}
  \propto p(\thetab) \prod_{i=1}^{N} \bigl[p(y_i \mid \thetab, \xib)\bigr]^{\nu_i}\,.
\end{equation}
This formulation represents a geometric mixture of the individual posteriors \(p(\thetab \mid y_i,\xib)\) and is used as an efficient importance sampling proposal when estimating gradients of the expected information gain. It aggregates information from possible outcomes of candidate experiment $\xib$.

\paragraph{\textbf{Gradient of Expected Information Gain}}
Using the pooled posterior, we can approximate the gradient of the expected information gain by Importance Sampling:
\begin{eqnarray}
    \nabla_\xib I(\xib) &=&
    \Exp_{p_\xib}\left[g(\xib,\Yb,\thetab,\thetab) -   \Exp_{q(\thetab' | \xib)} \left[\frac{p(\thetab' | \Yb, \xib)}{q(\thetab' | \xib)} \; g(\xib,\Yb,\thetab,\thetab')\right]  \right]\label{sharingan:EIG:gradRT2},
\end{eqnarray}

Using samples from $\thetab'_{j} \sim q(\thetab' | \xib)$ and $\thetab_{i} \sim p(\thetab)$ the gradient is estimated as:
    \begin{equation}
        \nabla_\xib I(\xib) \approx \frac{1}{N} \sum_{i=1}^N \left[ g(\xib,\yb_i, \thetab_{i},\thetab_{i})  - \frac{1}{M} \sum_{j=1}^{M} w_{i, j}\;  g(\xib,\yb_i, \thetab_{i}, \thetab'_{j}) \right] := \hat{\Gamma}(\thetab_{1:N}, \thetab'_{1:M}, \xib) \label{sharingan:expGMC}
    \end{equation}

With $g(\xib,\yb,\thetab, \thetab') = \nabla_\xib \log p(T_{\xib,\thetab}(\ub)| \thetab', \xib)_{|\ub = T^{-1}_{\xib,\thetab}(\yb)} \; $

\paragraph{\textbf{Sampling Operators}}\label{sharingan:sec:sampling_operators}
    From the reverse SDE \eqref{sharingan:eq:vp_reverse_sde} it is possible to derive an iterative scheme for sampling from the posterior distribution. For example, a simple Euler-Maruyama scheme takes the form:
    \begin{equation}\label{sharingan:eq:euler_maruyama_scheme}
        \thetab_{t-1} = \thetab_{t} + \left[-\frac{\beta(t_{t-1})}{2}\thetab_{t} - \beta(t_{t-1})\nabla\log p_t(\thetab_{t})\right]\Delta t + \sqrt{\beta(t_{t-1})\Delta t}\epsilon^{(t_{t-1})}
    \end{equation}

    An iterative sampling operator $\Sigma^{\Yb,\thetab}_t$ is defined such that \eqref{sharingan:eq:euler_maruyama_scheme} can be written as:
    \begin{equation}
        \thetab_{t-1} = \Sigma^{\Yb,\thetab}_t(\thetab_{t},\xib)
    \end{equation}

    Similarly, the pooled posterior is sampled following the backward SDE:
    \begin{equation}
        d\thetab^{'(t)} = \left[ - \frac{\beta(t)}{2} \thetab^{'(t)} - \beta(t)  \sum_{i=1}^N \nu_i \nabla_\thetab \log p(\thetab^{'(t)} |\yb_i^{(t)} , \xib)  \right] dt + \sqrt{\beta(t)} d\Bb_t \; .
        \label{sharingan:eq:diffexppost}
    \end{equation}
    After choosing a Conditional Diffusion sampling scheme to approximate the intractable $\log p(\yb_i^{(t)} | \thetab^{'(t)} , \xib)$ an iterative scheme is derived from \eqref{sharingan:eq:diffexppost} which using the sampling operator $\Sigma_t^{\thetab'}(q^{(t)},\xib,\rho)$ can be written as:
    \begin{equation}
        q^{(t-1)} = \Sigma_t^{\thetab'}(q^{(t)},\xib,\rho)
    \end{equation}

\section{Software details}
Our code is implemented in Jax \citep{jax2018github} and uses Flax as a Neural Network library and Optax as optimization one \citep{deepmind2020jax}.

\end{document}

%% file: preamble.tex
\usepackage[utf8]{inputenc} 
\usepackage[T1]{fontenc}    
\usepackage{hyperref}       
\hypersetup{
      colorlinks,
      urlcolor={black},
}
\usepackage{url}            
\usepackage{booktabs}       
\usepackage{amsfonts}       
\usepackage{nicefrac}       
\usepackage{microtype}      
\usepackage[table]{xcolor}         

\usepackage{subcaption}
\usepackage{multirow}
\usepackage[most]{tcolorbox}
\usepackage{lscape}

\usepackage{tabularx}
\usepackage{tikz,pgfplots}
\usepackage{tikz-3dplot}
\usetikzlibrary{decorations.text}
\usetikzlibrary{calc}

\usetikzlibrary{backgrounds,fit,decorations.pathreplacing, angles, quotes, patterns, math}  

\usepackage{graphicx}
\usepackage{amssymb,amsmath,enumerate}
\graphicspath{{pics/}{figs/}}
\usepackage{afterpage}
\usepackage{textcomp}

\usepackage{xspace}
\usepackage{bbm}

\usepackage[ruled,vlined]{algorithm2e}

\usepackage{pifont}

\def\XS{\xspace}
\DeclareMathAlphabet{\mathb}{OML}{cmm}{b}{it}
\def\sbm#1{\ensuremath{\mathb{#1}}}                
\def\sbmm#1{\ensuremath{\boldsymbol{#1}}}          

\def\Ab{{\sbm{A}}\XS}  
\def\Bb{{\sbm{B}}\XS}  
\def\Cb{{\sbm{C}}\XS}  
\def\Db{{\sbm{D}}\XS}  
\def\Eb{{\sbm{E}}\XS}  \def\eb{{\sbm{e}}\XS}
\def\Fb{{\sbm{F}}\XS}

\def\Ib{{\sbm{I}}\XS}

\def\Mb{{\sbm{M}}\XS}

\def\Sb{{\sbm{S}}\XS}  
  
  \def\ub{{\sbm{u}}\XS}

\def\Yb{{\sbm{Y}}\XS}  \def\yb{{\sbm{y}}\XS}

\def\epsilonb    {{\sbmm{\epsilon}}\XS}

\def\thetab      {{\sbmm{\theta}}\XS}      \def\Thetab    {{\sbmm{\Theta}}\XS}

\def\xib         {{\sbmm{\xi}}\XS}

\def\PM{\kern0pt^{\textrm{{\scriptsize PM}}}\kern0pt}
\def\MMAP{\kern1pt^{\textrm{{\tiny MMAP}}}\kern-1pt}

\def\Exp{\mathbb{E}} 
\newcommand{\Xv}{\ensuremath{\mathbf{X}}} 
\newcommand{\xv}{\ensuremath{\mathbf{x}}} 
\newcommand{\Yv}{\ensuremath{\mathbf{Y}}} 
\newcommand{\Zv}{\ensuremath{\mathbf{Z}}} 
\newcommand{\yv}{\ensuremath{\mathbf{y}}} 
\newcommand{\zv}{\ensuremath{\mathbf{z}}} 
\newcommand{\kv}{\ensuremath{\mathbf{k}}} 

\newcommand{\zero}{\ensuremath{\mathbf{0}}}

\def\Rset{\mathbb{R}} 
\def\wp1{\mathrm{w.p.} 1}  

\usepackage{amsthm}

\theoremstyle{plain}

